\documentclass[a4paper,12pt]{article}
\usepackage{amsmath}
\author{Oscar Stiffelman\\
    \texttt{ozzie@cs.stanford.edu}\\
    \texttt{imagine@gmail.com}
}

\title{The Least Wrong Model Is Not in the Data}
\DeclareMathOperator\len{len}

\begin{document}

\maketitle

\begin{abstract}
The true process that generated data cannot be determined when
multiple explanations are possible.  Prediction requires a model of
the probability that a process, chosen randomly from the set of
candidate explanations, generates some future observation.  The best
model includes all of the information contained in the minimal
description of the data that is not contained in the data.  It is
closely related to the Halting Problem and is logarithmic in the size
of the data.  Prediction is difficult because the ideal model is not
computable, and the best computable model is not ``findable.''
However, the error from any approximation can be bounded by the size of
the description using the model.

\end{abstract}

\subsection*{Introduction}

It is impossible to determine the true process that generated data
when multiple explanations are possible.  Each candidate process could
generate different observations in the future, so predicting
observations requires more than selecting a single best explanation.
A model is needed that can characterize the distribution of potential
observations generated by the candidate explanations.  If one only
considers computable processes, then the Church-Turing Hypothesis
tells us that each such process can be described by a Turing machine
which is specified by a binary string.  If an incorrect Turing machine
is used in place of the true process, then the data that the two
machines generate will eventually disagree.  A measurement of the
disagreement or error is the minimal description required to convert a
prediction into a future observation.  That difference is bounded by
the complexity required to convert the binary string specifying the
incorrect Turing machine into the string specifying the correct Turing
machine.  The Turing machine which is the minimal description of the
initial observations has the lowest average error in that it requires
the least complexity, on average, to convert from it to any other
candidate process.  However, it is not a model because it only
describes the already observed data, not the distribution of potential
observations.  The ideal model gives the probability that a process,
chosen at random from the set of candidate explanations, generates a
potential observation.  No computable function can determine this
exactly, but it will be shown that the best computable model is the
smallest prefix of a minimal description of the data that contains all
of the information that is not contained in the observed data.  The
remaining information in the minimal description was present in the
observed data and must be random.

\subsection*{Preliminary Definitions}

Let the observed sequence of data be represented by $x$, with $\len(x)$
indicating the number of bits in the representation.  The Kolmogorov
Complexity of $x$, indicated by $K(x)$ is the length of the shortest
bitstring defining a Turing machine that, when given to a reference
universal Turing machine, executes and halts leaving only $x$ on the
output tape~\cite{LiV08}.  It is a measurement of the intrinsic complexity of $x$.
If the data exhibits any regularity or non-independence between
different portions of the string, then it is compressible and $K(x) <
\len(x)$.  If the data is random, then the data is its own shortest
description.  The minimal program that generates $x$, or the first in
lexical order if there are multiple such programs, is indicated by
$x*$, with $\len(x*) = K(x)$. The conditional Kolmogorov Complexity of
a string, indicated by $K(a|b)$ is the length of the shortest program that
generates string $a$, possibly using the information in $b$.  The
programs are all assumed to be ``prefix-free'' in that no program is a
prefix of any other program, and each complexity is defined only to
within an O(1) factor to account for variation across different
universal Turing machines.  As a consequence of the Halting Problem,
the Kolmogorov Complexity is not computable.  The shortest program
that halts after generating some string cannot be determined by any
program because the set of programs that halt cannot be defined.

The set of all Turing machines that could have generated the data is
indicated by $Y = \{y_i\}$.  Each machine $y$ executes and halts after
generating $x$ and possibly an additional, subsequent string $w$.  The
assumption is that only $x$ has been observed at model selection time,
and that $w$ is not yet available.  If multiple symbols are generated
by a program, then some delimiter is used to indicate the end of
each symbol.

\subsection*{Minimal Expected Error} 
If an incorrect program from $Y$ is used instead of the true program
that generated the data, then the two programs will eventually
disagree.  There are many ways to define the error due to the
incorrect selection.  If we restrict ourselves to symmetrical error
functions that satisfy the triangle inequality, then the error can
also be interpreted as a distance function.  The length of the
shortest program that can generate either choice from the other is
minimal among all non-degenerate distance
functions~\cite[p.641]{LiV08}.  Using this definition of error, if $y$
is the correct program and $z$ is the incorrect program, then it can
be shown that the error is given by $$error(y,z) =
max(K(y|z),K(z|y)).$$

In order to identify the program that minimizes the expected error, it
is necessary to choose a specific probability distribution over the
programs in $Y$.  The universal probability distribution
$p(\alpha)=2^{-K(\alpha)}$ introduced by Ray Solomonoff approximates
\textit{any} computable measure $\mu(\alpha)$~\cite[p.348]{LiV08}\cite{solomonoff}.  Combining
the universal probability with the error definition, the expected
error is given by $$\sum_{y \in Y}\frac{max(K(y|z),K(z|y))}{2^{K(y)}}.$$

Because each program $y \in Y$ generates $x$, the conditional
complexity $K(x|y) = 0.$ The mutual information, or the information
contained in both strings, is $K(x) - K(x|y).$ This version of mutual
information is only symmetric to within a logarithmic factor, but a
related version that conditions on the minimal description is
symmetric: $K(x) - K(x|y*) = K(y) - k(y|x*)$~\cite[p.247]{LiV08}.
Because $y*$ includes all of the information in $y$, $K(x|y*)$ is also
equal to zero, and we can write $K(x) = K(y) - K(y|x*).$ Rearranging,
we can write $$K(y) = K(y|x*) + K(x).$$

This says that the conditional complexity required to generate each $y$
from $x*$ is the additional complexity contained in $y$ beyond that
contained in $x*$.  Because each $y$ generates $x$, the minimal
description of $y$ must contain the minimal description of $x$.  Using
the mutual information, the expected error expands to 
$$\sum_{y \in Y}{\frac{max(K(y|x*,z) + K({x*}|z), K(z|y))}{2^{K(y|x*)+K(x)}}}$$

When $z$ includes the information in $x*$, the $K({x*}|z)$ term goes
to zero for every item in the sum.  Adding additional information to
$z$ can decrease some of the $K(y|x*,z)$ terms, but it also increases
the $K(z|y)$ terms when the added information is not contained in $y$.
If the programs in $Y$ are independent, then adding more information
beyond $x*$ to $z$ can only reduce the error for an exponentially
small fraction of the programs, each weighted by the universal
prior, while increasing the error for the remaining programs.  The
expected error is therefore minimized by $x*$, the minimal description
of the data.  With that choice, the expected error becomes
$$
 \sum_{y \in Y}{\frac{max(K(y|x*),K({x*}|y))}{2^{K(y|x*) +K(x)}}} 
.$$

\subsection*{Ideal Model} 
Although the minimal description of the data minimizes the expected
error, it is not a model because it does not characterize the
distribution of potential observations.  The model needs to estimate
the conditional probability $p(w|x)$ of observing a given string $w$
after $x$, which is given by the ratio $$p(w|x) = \frac{p(xw)}{p(x)}.$$  It is possible
to again use Solomonoff's universal prior in place of the unknown
probability distribution~\cite[p.350]{LiV08}.  However, a perhaps more intuitive argument
is to assume that programs are generated entirely randomly by, for
example, flipping an unbiased coin~\cite{chaitin}.  Each coin flip generates a bit
that can be appended to a tape.  Using the prefix-free assumption, any
extra bits can be ignored.  Even though there is no complexity prior
or selection bias, simple programs are exponentially more likely to be
drawn by this process.  If $K(\alpha)$ is the minimal number of bits
required to describe some string $\alpha$, then the fraction of
random strings of length $n$ that start with that minimal description is
$$\frac{2^{n-K(\alpha)}}{2^n} = 2^{-K(\alpha)}.$$ The total fraction
of strings that generate $\alpha$ may be larger than that because it
includes non-minimal descriptions of $\alpha$.  However, the
non-minimal descriptions are more complex and, therefore,
exponentially less frequent than the minimal description.  Therefore,
the probability of selecting a program at random that generates
$\alpha$ is approximately $$p(\alpha) = 2^{-K(\alpha)}.$$

Using this same argument, the conditional probability is given by 
\begin{align*}
p(\beta|\alpha) &= \frac{p(\alpha\beta)}{p(\alpha)} \\
                &= \frac{2^{-K(\alpha,\beta)}}{2^{-K(\alpha)}}.
\end{align*}

Using the symmetric mutual information, the term $K(\alpha,\beta)$
expands to $K(\alpha) + K(\beta|\alpha*).$  The conditional
probability of observing some string $w$ after observing the data $x$
is therefore given by
\begin{align*}
 p(w|x) &= \frac{2^{-K(x,w)}}{2^{-K(x)}} \\
        &= \frac{2^{-K(x) - K(w|x*)}}{2^{-K(x)}} \\
        &= 2^{-K(w|x*)}
\end{align*}

The ideal model predicts future observations using $p(w|x) =
2^{-K(w|x*)}.$  However, it is not a computable function because the
conditional complexity $K(w|x*)$ is not computable.

\subsection*{Computable Model}
The ideal model determines the distribution of potential observations
generated by candidate programs.  That distribution is defined by the
length of the shortest program that generates the predicted
observation from the minimal description of the observed data:
$K({w}|x*)$.  No computable function can always find the shortest
program because the set of halting programs cannot be defined.
However, if some program can be found that generates $w$ from $x*$,
then the length of that program is an upper bound on the length of the
minimal program.  Let $h(w)$ indicate the best computable model.  If
the computable model determines the length of some program that
generates $w$ from $x*$, then it must be an upper approximation:
$h(w) >= K({w}|x*)$.

The model function $h(w)$ returns an integer that is an upper approximation
of $K({w}|x*)$.  The best choice of $h(w)$ is difficult to characterize
in this initial and general formulation.  However, it is possible to
define some additional, yet equivalent, functions for which the
optimal solution is apparent.  For any choice of $h$, a function $g$
can be defined that returns strings that are of length equal to the
value returned by $h$: $g(w) = r_w$, and $\len(r_w) = h(w)$.  Because
there can be no more than $2^{h(w)}$ programs of a given length
$h(w)$, it is always possible to uniquely assign the strings $r_w$
returned by $g(w)$ while preserving the correspondence $\len(r_w) =
h(w)$.  When the function $g(w)$ is unique, it is invertible.  Let the
inverse of $g$ be indicated by $f(r_w) = w$.  The function $g$ is
computable because the function $h$ is computable.  This means that
$g(w)$ can be expressed as a Turing machine: $[g\;w]\to r_w$.  Because
$f(r_w)$ is the inverse of $g$, it is also computable and can be
expressed as a Turing machine: $[f\;r_w]\to w$.  For any computable
function, the output can be fully determined from the function and the input
parameter, so the conditional complexity of the output given the input
is zero: $K(r_w|g,w) = 0$ and $K(w|f,r_w) = 0$.  Because the inverse
of a computable function can be determined from the function by
iterating through the input space until a given value is generated,
the conditional complexity of the inverse function given the initial
function is zero.  The statement $K(r_w|f,w) = 0$ follows from the
combination of $K(r_w|g,w) = 0$ with $K(g|f)=0$.  Here is a summary of
the functions and relations.
\begin{align*}
h(w) &\geq K({w}|x*) \\
g(w) &= r_w \\
f(r_w) &= w \\
f &= g^{-1} \\
\len(r_w) &= h(w) \\
K(r_w|f,w)&=0
\end{align*}

An object is random when it is incompressible and therefore is its own
shortest description: $K(\gamma) = \len(\gamma)$.  If an object is
compressible, then it must exhibit some regularity or
non-randomness. If that regularity can be specified, then the
remaining information needed to describe the object is incompressible
and the object can be considered conditionally random with respect to
the description of the regularity.  Just as the minimal information
needed to fully describe a random object is contained in the object,
so too is the remaining information needed to describe a conditionally
random object fully contained in the object.  If an object $\gamma$ is
conditionally random with $\alpha$, then there is some minimal string
$\beta$ of length $K(\gamma | \alpha)$ that can be determined from
$\alpha$ and $\gamma$, and that can be combined with $\alpha$ to
generate $\gamma$.  Because $\beta$ is determined from $\alpha$ and
$\gamma$, its conditional complexity is $K(\beta|\alpha,\gamma) = 0$.
If $\gamma$ is not random with $\alpha$, then any string $\beta$ that
is determined from the data, and for which $[\alpha \beta]$ generates
$\gamma$ must be larger than the minimal $K(\gamma|\alpha)$.  If
$\alpha$ is the smallest string for which $\gamma$ is conditionally
random, then $\gamma$ is maximally conditionally random with $\alpha$,
and $\len(\alpha) + \len(\beta) = K(\gamma)$.

It will now be shown that the computable model $h(w)$ is a best
approximation for the ideal model when $x$ is maximally random with
$f$.  For notational convenience, let $\gamma \in
\hat{K}(\alpha|\beta)$ indicate that $\gamma$ is a program of length
$K(\alpha|\beta)$ that generates $\alpha$ from $\beta$.  Any string
$x*$ can be factored into two parts $[a\; b]$ where $a \in
\hat{K}({x*}|x)$ and $b \in \hat{K}({x*}|a)$.  Because the complete
$x*$ can be generated from $x$ and $a \in \hat{K}({x*}|x)$, and
because $b$ is a component of $x*$, it is clear that $K(b|a,x) = 0$
and $K(x|a) = b$.  Therefore $x$ is conditionally random with
$a \in \hat{K}({x*}|x)$.  It is also clear that $K({x*}|x)$ is the length of the
smallest string for which $x$ is conditionally random.  By definition,
it is the information that is needed to describe $x*$ that is not in
$x$, so if any information is removed, any subsequent choice of $b$
determined from the data would have to exceed the minimal description.
Therefore $x$ is maximally conditionally random with $a \in
\hat{K}({x*}|x)$.

When $x$ is maximally random with $f$, if $w$ is also maximally random
with $f$, then $$K({w}|x*) = K(f,r_w|f,r_x) = K(r_w|r_x) \leq
\len(r_w).$$  With the exception of an exponentially small fraction of
the strings, the $r_w$ and $r_x$ are independent of each other, and
$K({w}|x*) = \len(r_w)$.  The computable model's approximation of the
distribution is given by $h(w) = \len(r_w)$, and the
ideal distribution is equal to $K({w}|x*)$.  Therefore the computable
model exactly matches the ideal model for almost all strings that are
maximally random with $f$.  Any increase in $f$ to reduce $r_w$ in the
small fraction of cases where $K(r_w|r_x) < K(r_w)$ would lead to an
increase in the $r_w$ for the rest of the maximally random strings in
the set because the $r_w$ must satisfy the Kraft Inequality $\sum
2^{-\len(r_w)} \leq 1$.  Any decrease in $f$ would increase all $r_w$ in the
set because each $w$ would no longer be random with $f$ and so the
$r_w$ would have to increase by even more than the $f$ decreased
because $r_w$ would have to grow larger than the conditional
complexity.  Therefore, the choice of $f$ is optimal for the set of
strings that are maximally random with $f$.

In fact, the optimality extends to all $w$ that are random
with $f$, even if they are not maximally random.  Using the symmetric
mutual information, the fact that $[f \;r_w]$ generates $w$ and is
incompressible means that if $w$ is random, but not maximally random,
with $f$, then a portion of $w$ is included in $f$, and the rest is
defined by $r_w$: 
\begin{gather*}
K(w|f,r_w) = 0 \\
K(w) - K(w|f,r_w) = K(w) = K(f,r_w) - K(f,r_w|{w}) \\
K(f,r_w) - K(w) = \len(f) + \len(r_w) - \len(w) = K(f,r_w|{w})
\end{gather*}
Therefore $K({w}|x*) = K(f,r_w|f,r_x) = K(r_w|r_x) \leq \len(r_w)$, as
was the case for strings that were maximally random with $f$.

When strings are random with $f$, the approximation is optimal.  When
strings are not random with $f$, the approximation defined by $r_w$
will overestimate the conditional complexity $K({w}|x*)$.  Because
most strings are incompressible and independent of $x$, the best
estimate for $K({w}|x*)$ in those cases is simply $\len(w)$.  Even if
strings are compressible, there is no computable function that can be
expected to identify the regularity in general.  The only regularity
that can be exploited by $f$ is information that has already been
identified in $x*$.  Therefore, the optimality of the approximation
for the remaining strings depends on how closely $r_w$ matches
$\len(w)$.  Because $f(r_w)$ is computable, it corresponds to a set of
prefix-free programs that begin with $f$.  The strings $r_w$ form a
prefix-free code tree.  Any code paths that are not branching can be
compacted to create an equivalent yet shorter codeword.  If the
function $f(r_w)$ is also restricted so that each string $r_w$
corresponds to a unique $w$, then the set of codewords is
incompressible.  Given these assumptions, no $r_w$ can be much larger
than $\len(w)$ because $\len(r_w) = K(r_w)$, and $K(r_w|f,w) = 0$.  The
approximation of the computable model is therefore optimal for strings
that are random with $f$, and not much worse than any computable
function for the remaining strings.

\subsection*{Model Size and the Halting Problem}

The optimal computable model, which is the best approximation to the
distribution of observations, is defined by a program prefix of size
$K({x*}|x)$.  This is the information needed to construct the minimal
description of the data that is not available from the data.  The
Kolmogorov Complexity is an undecidable function.  It is not possible
to run every Turing machine in order of size and select the smallest
one that halts after generating the data because no program can
determine which other programs will halt.  However, if the size of the
minimal description is known, a minimal description can be found by
running all programs of that size in parallel and selecting the first
one that halts after generating the data~\cite[p.~252]{LiV08}.  Because the minimal
description is no larger than the data, the size of the minimal
description can be encoded using $\log(\len(x))$ bits when $\len(x)$ is
available.  The size of the optimal model is therefore logarithmic
with the data: $$K({x*}|x) <= \log(\len(x)).$$

\subsection*{Kolmogorov's Randomness Deficiency}
Kolmogorov proposed (but never published) an approach to
``nonprobabilistic statistics'' that is closely related to this work~\cite{structure}.
In his approach, models are defined as Turing machines that enumerate
finite sets of objects.  Given a model set $S$, the index of the
object within the set can be encoded using $\log|S|$ bits, referred to
as a data-to-model code, and this provides one way to describe the
object.  The difference between $\log|S|$ and the conditional
complexity is defined as the randomness deficiency: $\log(|S|) -
K(x|S)$.  When that quantity is small, the object is random or typical
with respect to the set and there are no simple properties that
distinguish the object from the majority of elements in the set.  The
best fit model at any complexity level is the model with the lowest
randomness deficiency, and the least complex model achieving $K(S) +
\log(|S|) = K(x)$ describes all of the regularity of the data.
Although the randomness deficiency cannot be approximated and is
therefore not a suitable criteria for model selection, it can be shown
that a model with minimal $\log(|S|)$ or $\log(|S|) + \alpha$ among all
candidate models of complexity less than $\alpha$ also achieves
minimal randomness deficiency.  Because the data-to-model code is
computable, unlike the randomness deficiency, this can be used in an
effective procedure for model selection.

In this work, the best model determines the distribution of potential
strings that could be generated by the set of candidate programs.  The
best computable model is the best computable approximation of that
distribution.  It is interesting to compare the best model according
to this criteria to the best model according to Kolmogorov's
Randomness Deficiency criteria.  Although sets of objects are not
explicitly enumerated by the models in this work, the $r_w$ component
is analogous to the data-to-model code in Kolmogorov's Randomness
Deficiency approach.  For a given inverse model function $f(r_w) = w$,
a Turing machine could be constructed that enumerates the set of $w$
that can be generated from codewords of a given length.  The
distribution is optimally approximated when $x$ is maximally random
with the inverse model function $f$.  When $x$ is random with $f$,
$\len(r_x) = K(x|f)$.  And when it is maximally random, $f$ is the
smallest function for which that is true.  Therefore $f$ also
minimizes the randomness deficiency, and the best estimate of the
distribution is also a best model according to Kolmogorov's Randomness
Deficiency criteria.

\subsection*{Approximation, MDL, and Hillclimbing}
The best computable model may not be in the class of candidate models.
Even if it is in the class, there is no way to know when it has been
selected because it can be used to construct the minimal description
of the data and no computable function can confirm that the minimal
description has been selected.  Although no strategy for selecting the
best computable model can be devised, it is possible to bound the
error from any model approximation. That bound can be used to select a
model which is the closest approximation to the best computable model.

The best computable model is defined by the inverse model function
$f(r_w)=w$.  If a candidate model is described by
$\tilde{f}(\tilde{r}_w) = w$, then the error is at least as large as
$K(r_w|\tilde{r}_w)$ because that is the minimal complexity required
to generate the correct $r_w$ from the approximate $\tilde{r}_w$.
Because $r_w$ is fully determined from $f$ and $x$, the error is no
less than $K(f|\tilde{f})$.  The best approximation in the model class
is the model with minimal $K(f|\tilde{f})$.

To simplify notation, let $A=fr_x$, and let $B=\tilde{f}\tilde{r}_x$.
Using the symmetric mutual information, $K(A) - K(A|B*) = K(B) -
K(B|A*)$.  Because $B$ generates $x$, $K(x) - K(x|B*) = K(x) = K(B) -
K(B|x*)$.  The minimal description of $x$ is given by $A$, so $A=x*$,
and $A* = A$.  Therefore $K(B|A*) = K(B)-K(A).$ Combining, this gives
\begin{align*}
K(A) - K(A|B*) &= K(B) - K(B|A*) \\
K(A) - K(B) + K(B|A*) &= K(A|B*) \\
K(A) - K(B) + (K(B)-K(A)) &= K(A|B*)\\
K(A|B*) &= 0 \\
K(A|B) &= K(A|B*) + K({B*}|B)  \\
K(A|B) &= K({B*}|B) \leq \log(\len(B)) \\
\end{align*}

The last inequality follows from the fact that the minimal description
of a string can be generated from the length of the minimal
description and the string.  This expands to $$K(f,r_x |
\tilde{f},\tilde{r}_x) <= \log(\len(\tilde{f}) +
\len(\tilde{r}_x)).$$  If $\tilde{r}_x$ is the non-compressive
component of the candidate model, then it shares no mutual
information with $f$, which only describes the compressive regularity
of the data.  Assuming that constraint, the expression expands
to
\begin{align*}
K(f|\tilde{f}) + K(r_x|\tilde{f},\tilde{r}_x) &\leq
\log(\len(\tilde{f}) 
+ \len(\tilde{r}_x)) \\
K(f|\tilde{f}) &\leq \log(\len(\tilde{f})
+ \len(\tilde{r}_x)) 
\end{align*}

The error is bounded by the sum of the model description and the data
description using the model.  Minimizing the description length is
therefore an effective strategy for selecting the best approximation.
This is the strategy advocated by the minimum description length (MDL)
principle~\cite{MDL}.  However, the error bound depends on the
assumption that the model is the only component of the description
that is compressive.  Minimizing the description length without that
constraint does not minimize the error.

Another caveat is that the model description may overestimate the
model complexity to such an extent that it does not correctly identify
the best approximation.  In practice, models can be very high
dimensional.  Starting from an initial, poorly performing model,
gradient descent or some other hillclimbing strategy is typically
applied to find a model that performs better on training data.
However, hillclimbing can only be effective if there is mutual
information between nearby points in the model space.  In general
terms, this means that the more amenable a class of models is to
optimization, the more poorly its description length serves as a
criteria for model selection.  When description length is close to
model complexity, the class of models is incompressible, and optimization
is no better than exhaustive enumeration.

\subsection*{Probability Models}
There is a direct equivalence between computable models and
probability distributions.  Every computable model defines a
probability distribution, and every computable probability
distribution is defined by a computable model.  The inverse model
functions $f(r_w)=w$ are constructed by prefix-free Turing machines,
so the set of $r_w$ codewords is also prefix-free.  The Kraft
Inequality guarantees that the codewords can be interpreted as a
probability distribution: $\sum 2^{-\len(r_w)} <= 1$.  That distribution is
the best computable approximation to the distribution of future
observations, so it is reasonable to interpret $2^{-\len(r_w)}$ as the
probability of observing the string $w$.  And every computable probability
distribution can be expressed as a set of prefix-free codewords
associated with some inverse model function.

Although the two kinds of models are fundamentally equivalent,
classical probability models are generally defined over a restricted,
finite set of symbols, whereas Turing machines can construct
arbitrarily long strings.  A string that does not exhibit any
frequency regularity for the predefined set of symbols cannot be
compressed by a restricted probability model.  However, there may be
regularity in the data that can be described by some program.  For
example, a sequence of coin flips that alternate perpetually (HTHTHT
...) exhibits predictable regularity~\cite[P.48]{LiV08}.  But the
frequency of each symbol is 1/2 and the sample entropy is 1, so the
restricted probability model would require N bits to encode N coin
flips.  Restricting the class of models may exclude the optimal model.
As with any restricted class of models, the error is minimized by
choosing the model for which the sum of the model complexity and the
remaining, non-compressive data complexity is minimal.  In the case of
probability models, that sum is given by $K(model)
+\log(1/{P(data)}).$

\subsection*{Entropy and Complexity}
Important and subtle relationships between model complexity and
entropy arise when considering probability models.  Entropy measures
the average information required to describe symbols from a
distribution, independent of the objects associated with those
symbols~\cite{Cover}.  In contrast, Kolmogorov Complexity characterizes the
information required to describe individual objects~\cite{LiV08}.  If data can be
encoded more compactly using a given probability model, relative to
the least complex uniform distribution, then $$ K(q) + \log(1/q(x)) <
\log(1/u(x)) = n * \log(m)$$ where $q$ is the model distribution, and $u$
is the uniform distribution that assigns equal probability to each of
the $m$ symbols.  If $p$ is the empirical distribution of the data
(the frequency of each predefined symbol in the data), then this can
be expressed as $K(q) < n * \sum p(i) * \log(q(i)/u(i))$.  The
complexity of the model must not exceed the reduction in the
complexity due to the use of the model.  Simplifying, this becomes

\begin{align*}
K(q) &< n*\sum p(i)*\log(1/u(i)) - n*\sum p(i)*\log(1/q(i)) \\
 &< n * \log(m) - n * \sum p(i)*\log(1/q(i))  \\
 &< n * \log(m) -  n * \sum p(i)*\log(1/p(i)) \\
 &< n * (H(u) - H(p)) \\
\end{align*}
where H is the entropy of the distribution, and the Gibb's Inequality
has been used to substitute $p$ for $q$ in the third
inequality~\cite{Cover}.  The quantity $n * (H(u) - H(p))$ is an upper
bound on the complexity of any viable probability model and is equal
to the Kullback-Leibler divergence between the uniform and empirical
distributions, scaled by the amount of data.  When the entropy of the
data is high, the complexity of the model must be low.  The complexity
of the model can increase only as the empirical distribution diverges
from the uniform distribution.

As the model distribution approaches the
empirical distribution, it is reasonable to view the empirical
distribution also as an approximation of the model distribution.
Although not strictly true, this suggests a roughly inverse
relationship between the entropy and the complexity of the model.
However, the intuition only applies to selected models, not to
candidate models.  Jaynes' Maximum Entropy model selection criteria
implies that entropy and model complexity should have a more
fundamental relationship~\cite{Jaynes2003}.  That relationship is made
clear by considering the two-part description of the Turing machine
for the model.  The total complexity of each symbol generated by the
model is $$K(q) + \log(1/q(\alpha)) >= K(q) + K(\alpha|q) >= K(\alpha).$$  The model
complexity must always satisfy $$K(q) >= K(\alpha) - \log(1/q(\alpha)),$$ so the
model complexity may need to increase if the codeword decreases.  A
distribution with maximum entropy has a uniform codeword for each
symbol.  As the entropy of the distribution decreases, the codewords
for some of the symbols become small compared to others.  If the
symbols are of approximately equal complexity, the difference in the
complexity must be shifted into the model.  A more general
interpretation is that if the entropy or uncertainty required to
describe a system decreases, then either some compressible structure
has been identified, or the information has been shifted into the
model.  As most data is not compressible, lower entropy descriptions
are generally associated with higher complexity models.

\subsection*{Conclusion}
This paper analyzes the problem of prediction from first principles.
The ideal but uncomputable model, the best computable but
``unfindable'' model, and the best findable approximating model are
each characterized.  There is no way to identify the true, hidden
process that generated data.  However, the probability of any
prediction can be defined as the probability of a randomly chosen
candidate process generating it.  That probability is equal to the
uncomputable function $2^{-K({w}|x*)}$, where $K({w}|x*)$ is the
shortest description of the prediction given the shortest description
of the observed data.  The best computable approximation to that
uncomputable function is a model constructed from a string of length
$K({x*}|x)$ containing all of the regularity of the data that cannot
be determined from the data.  The best model is closely related to the
Halting Problem.  The undecidability of the Halting Problem is the
reason that the information could not be determined from the data.
The size of the best computable model is logarithmic with the data
based on its relationship to the Halting Problem.  Although the best
model is computable, it is ``unfindable'' because it includes the
uncomputable complexity of the data.  Even if the correct model has
been selected, there is no way to know that it has been selected.
Although the best model cannot be found, the error from using an
approximation is bounded by the total description length of the data
using the approximating model.  The optimal computable model is also a
best fit model by Kolmogorov's Randomness Deficiency criteria, and the
best approximate model is consistent with the MDL principle and the
Maximum Entropy Principle in the context of probability models.

\subsection*{Acknowledgements}
The author wishes to acknowledge the valuable discussions and feedback
provided by Hein Roehrig.  Correspondence with Paul Vit\'anyi about
Kolmogorov's structure function, lossy compression, and denoising,
was also very helpful.

\bibliography{bestmodel}
\bibliographystyle{alpha}

\end{document}